\documentclass{article}

\usepackage{spconf}

\usepackage[xetex]{graphicx}
\usepackage{import}  
\usepackage{tikz}  
\usepackage{calc}  
\usepackage{arydshln}  
\graphicspath{{fig/}}

\newcommand{\mc}[1]{\multicolumn{1}{c}{#1}}  
\usepackage{multirow}
\usepackage{booktabs}

\usepackage{microtype}
\def\mytodoenabled{}
\usepackage[prependcaption,textsize=scriptsize\mytodoenabled]{todonotes}
\makeatletter  
\def\myenabledtest{}
\newcommand{\todoref}[1][\@nil]{[
	\ifx\mytodoenabled\myenabledtest
		\def\refarg{#1}
		\ifx\refarg\@nnil
			{\scriptsize\textcolor{red}{(TODO) REF}\PackageWarning{TODO}{(TODO) REF}}
		\else
			{\scriptsize\textcolor{red}{(TODO) REF: #1}\PackageWarning{TODO}{(TODO) REF: #1}}
		\fi
	\fi
]}
\makeatother  
\newcommand\blfootnote[1]{\begingroup
                          \renewcommand\thefootnote{}\footnote{#1}
                          \addtocounter{footnote}{-1}
                          \endgroup}

\newcommand{\hk}[1]{#1}

\usepackage{cite}
\let\oldbibliography\thebibliography
\renewcommand{\thebibliography}[1]{\oldbibliography{#1}
                                   \setlength{\itemsep}{0.4pt}
                                   \vspace*{-0mm}}

\usepackage{url}
                                   
\usepackage{amsmath,amsfonts,bm}

\def\vx{{\mathbf{x}}}

\title{Multimodal One-Shot Learning of Speech and Images}

\name{
	Ryan Eloff \qquad
	Herman A. Engelbrecht \qquad
	Herman Kamper\vspace*{0pt}
}

\address{
        E\&E Engineering, Stellenbosch University, South Africa \\
	{\small \tt rpeloff@sun.ac.za, hebrecht@sun.ac.za, 
	kamperh@sun.ac.za}\vspace*{0pt}
}

\begin{document}
\ninept
\maketitle

\begin{abstract}
Imagine a robot is shown new concepts visually together with spoken tags, e.g.\ ``milk'', ``eggs'', ``butter''. After seeing \textit{one} paired audio-visual example per class, it is shown a new set of unseen instances of these objects, and asked to pick the ``milk''. Without receiving any hard labels, could it learn to match the new continuous speech input to the correct visual instance? Although unimodal one-shot learning has been studied, where one labelled example in a single modality is given per class, this example motivates \textit{multimodal one-shot learning}. Our main contribution is to formally define this task, and to propose several baseline and advanced models. We use a dataset of paired spoken and visual digits to specifically investigate recent advances in Siamese convolutional neural networks. Our best Siamese model achieves twice the accuracy of a nearest neighbour model using pixel-distance over images and dynamic time warping over speech in 11-way cross-modal matching.
\end{abstract}



\section{Introduction}
\label{sec:intro}
Humans possess the remarkable ability to learn new words and object categories from 
only one or a few examples~\cite{Halberda2006Dax}.
\hk{For example, a child hearing the word ``lego'' for the first time in the context of receiving a new toy, can quickly learn to associate the spoken word ``lego'' to the new (visual) concept \textit{lego}.}
Current state-of-the-art speech and vision \hk{processing} algorithms require thousands of labelled examples to complete a similar task. 
This has lead to research in \textit{one-shot learning}~\cite{FeiFei2006OneShot, Lake2011OneShotVis, Koch2015Siamese}, where the task is acquisition of novel concepts from only one or a few labelled examples.

One-shot learning studies 
have focused on problems where novel concepts \hk{in a single modality} are observed along with class labels.
\hk{This is different from the example above:}
the child directly associates the spoken word \hk{``lego''} to the visual signal of lego without any \hk{class labels}, 
and can generalise this single example to other visual or spoken instances of \hk{\textit{lego}}.
This motivates \textit{multimodal one-shot learning}, a new task 
we formalise in this \hk{paper}. 
Consider an agent such as a household robot that is shown a single visual \hk{example} of \hk{\textit{milk}, \textit{eggs}, \textit{butter} and a \textit{mug}}, novel objects each paired with a spoken description.
\hk{During subsequent use, a speech query is given and the agent needs to identify which visual object the query refers to}.
This setting is relevant in \hk{modelling} infant language acquisition, \hk{where models can be used to test particular cognitive hypotheses}~\cite{Rasanen2015HumanLang};
low-resource speech processing, \hk{where new concepts could be taught in an arbitrary language}~\cite{Besacier2014LowResASR};
and robotics, where novel concepts must be acquired online from co-occurring multimodal sensory inputs~\cite{Walter2012OneShotRobot, Taniguchi2016Robotics, Thomason2016RobotGestures, Renkens2017SpeechAcqui, Renkens2018CapsNetSLU}.

Here we specifically consider multimodal one-shot learning on a dataset of isolated spoken digits paired with images.
A model is shown a set of speech-image pairs, one for each of the 11 digit classes. We refer to this set, which is acquired before the model is applied, as the \textit{support set}.
During testing, the model is shown a new instance of a spoken digit and a new set of test images, called the \textit{matching set}.
It then needs to predict which test image in the matching set corresponds to the spoken input query.
We tackle this problem by extending existing unimodal one-shot models to the multimodal~case. 

\hk{
To do cross-modal test-time matching, 
we propose a framework relying on 
unimodal comparisons through the support set.
Given an input speech query, we find the closest speech segment in the support set.
We then take its paired support image, and find its closest image in the matching set.
This image is predicted as the match.
Metrics for speech-speech and image-image comparisons need to be defined, and this is where we take advantage of the large body of work in unimodal one-shot learning to investigate several options.
One approach is to use labelled background training data not containing any of the classes under consideration.
Using such speech and image background data, we specifically investigate {Siamese neural networks}~\cite{Bromley1993Siamese, Chopra2005Siamese} as a way to explicitly train unimodal distance metrics.
We also incorporate recently proposed advances~\cite{Koch2015Siamese, Chechik2010OASIS, Wang2014TripNet, hermann2014multilingual, Hoffer2015TripletNet, Schroff2015FaceNet, Hermans2017TripletLoss} for such networks.
}

\hk{We}
compare novel Siamese convolutional neural network (CNN) architectures to traditional direct feature matching models. 
We show that a single CNN with a triplet loss~\cite{Chechik2010OASIS, Wang2014TripNet} and online semi-hard mining~\cite{Schroff2015FaceNet} is more efficient and results in higher accuracies 
than the offline variant \hk{which} uses shared weight networks\hk{, both approaches outperforming the direct feature matching baseline}.
Our \hk{main} contribution is 
the formal definition of multimodal one-shot learning.
\hk{We also develop a}
one-shot cross-modal matching \hk{dataset that} 
may be used to benchmark other 
approaches.
\hk{As an intermediate evaluation in our work, we also consider unimodal one-shot speech classification.}
Apart from~\cite{Lake2014OneShotSpeech}, \hk{our paper is} 
to our knowledge the only work that considers one-shot unimodal learning of spoken language.
We present several new models and baselines not considered in~\cite{Lake2014OneShotSpeech}.\footnote{Full code recipe available at: \url{https://github.com/rpeloff/multimodal-one-shot-learning}.}


\begin{figure*}[!h]
\centering
\scalebox{0.95}{
{\footnotesize (a)}
\sffamily  
\begin{tabular}{c:c}
	\begin{minipage}[b]{.145\linewidth}  
		\centering
	  	\centerline{
	  		\footnotesize
			\def\svgscale{0.23} 
			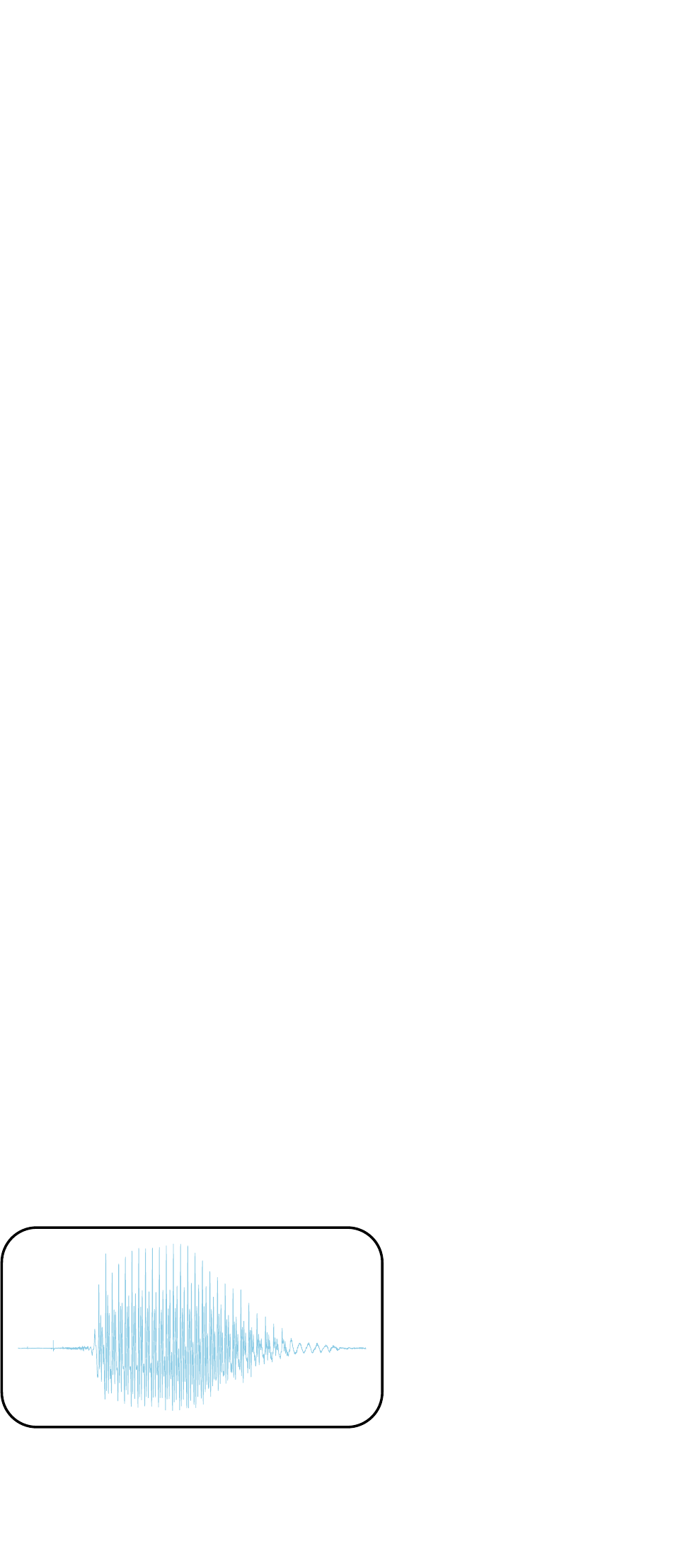
		}
		\vspace*{-0.2cm}
	\end{minipage}&
	\begin{minipage}[b]{.190\linewidth}  
		\centering
		\centerline{
			\footnotesize
			\def\svgscale{0.23}
\begingroup%
  \makeatletter%
  \providecommand\color[2][]{%
    \errmessage{(Inkscape) Color is used for the text in Inkscape, but the package 'color.sty' is not loaded}%
    \renewcommand\color[2][]{}%
  }%
  \providecommand\transparent[1]{%
    \errmessage{(Inkscape) Transparency is used (non-zero) for the text in Inkscape, but the package 'transparent.sty' is not loaded}%
    \renewcommand\transparent[1]{}%
  }%
  \providecommand\rotatebox[2]{#2}%
  \newcommand*\fsize{\dimexpr\f@size pt\relax}%
  \newcommand*\lineheight[1]{\fontsize{\fsize}{#1\fsize}\selectfont}%
  \ifx\svgwidth\undefined%
    \setlength{\unitlength}{425.19685039bp}%
    \ifx\svgscale\undefined%
      \relax%
    \else%
      \setlength{\unitlength}{\unitlength * \real{\svgscale}}%
    \fi%
  \else%
    \setlength{\unitlength}{\svgwidth}%
  \fi%
  \global\let\svgwidth\undefined%
  \global\let\svgscale\undefined%
  \makeatother%
  \begin{picture}(1,1.5)%
    \lineheight{1}%
    \setlength\tabcolsep{0pt}%
    \put(0,0){\includegraphics[width=\unitlength,page=1]{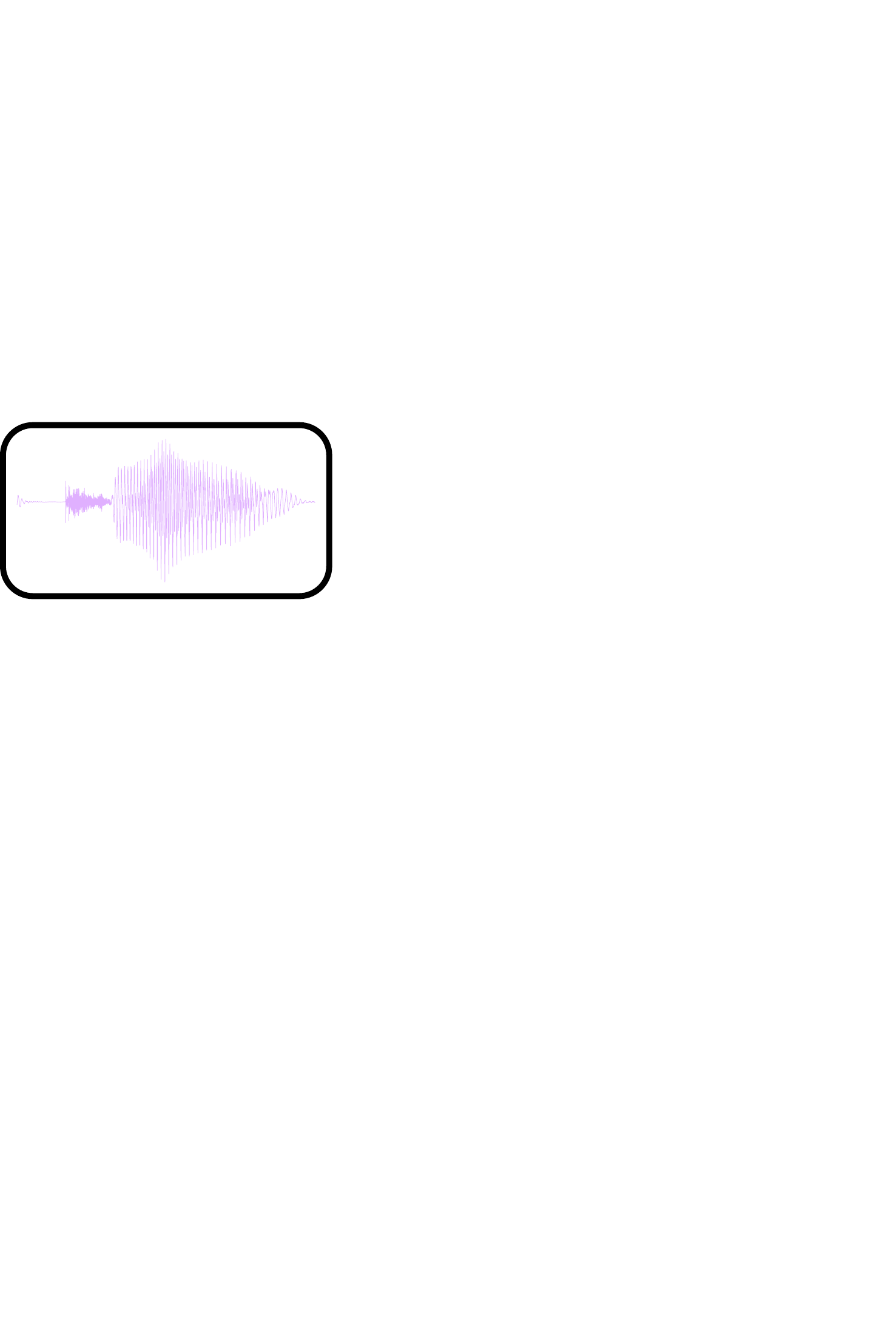}}%
    \put(0.31,1.075){\color[rgb]{0,0,0}\makebox(0,0)[lt]{\lineheight{1.25}\smash{\begin{tabular}[t]{l}\small{$C_{\mathcal{S}}(\hat{\vx}_a)$}\end{tabular}}}}%
    \put(-0.00158474,0.70460042){\color[rgb]{0,0,0}\makebox(0,0)[lt]{\lineheight{1.25}\smash{\begin{tabular}[t]{l}Query $\hat{\vx}_a$\\(\textit{``two''})\end{tabular}}}}%
    \put(0,0){\includegraphics[width=\unitlength,page=2]{unimodal_task_infer.pdf}}%
    \put(0.60152664,0.91){\color[rgb]{1,0,0}\makebox(0,0)[lt]{\lineheight{1.25}\smash{\begin{tabular}[t]{l}\textcolor{red}{$\hat{y}=$ ``two''}\end{tabular}}}}%
  \end{picture}%
\endgroup%

		}
		\vspace*{-0.2cm}
	\end{minipage}\\
	\mc{\vspace*{0.1cm}}&\mc{\vspace*{0.1cm}}\\
	\mc{\footnotesize \shortstack{i. One-shot speech \\ learning}}&
	\mc{\footnotesize \shortstack{ii. One-shot speech \\ classification}}
\end{tabular}
\rmfamily  
{\footnotesize (b)}
\sffamily  
\begin{tabular}{c:c}
	\begin{minipage}[b]{.210\linewidth}  
		\centering
		\centerline{
			\footnotesize
			\def\svgscale{0.23}
			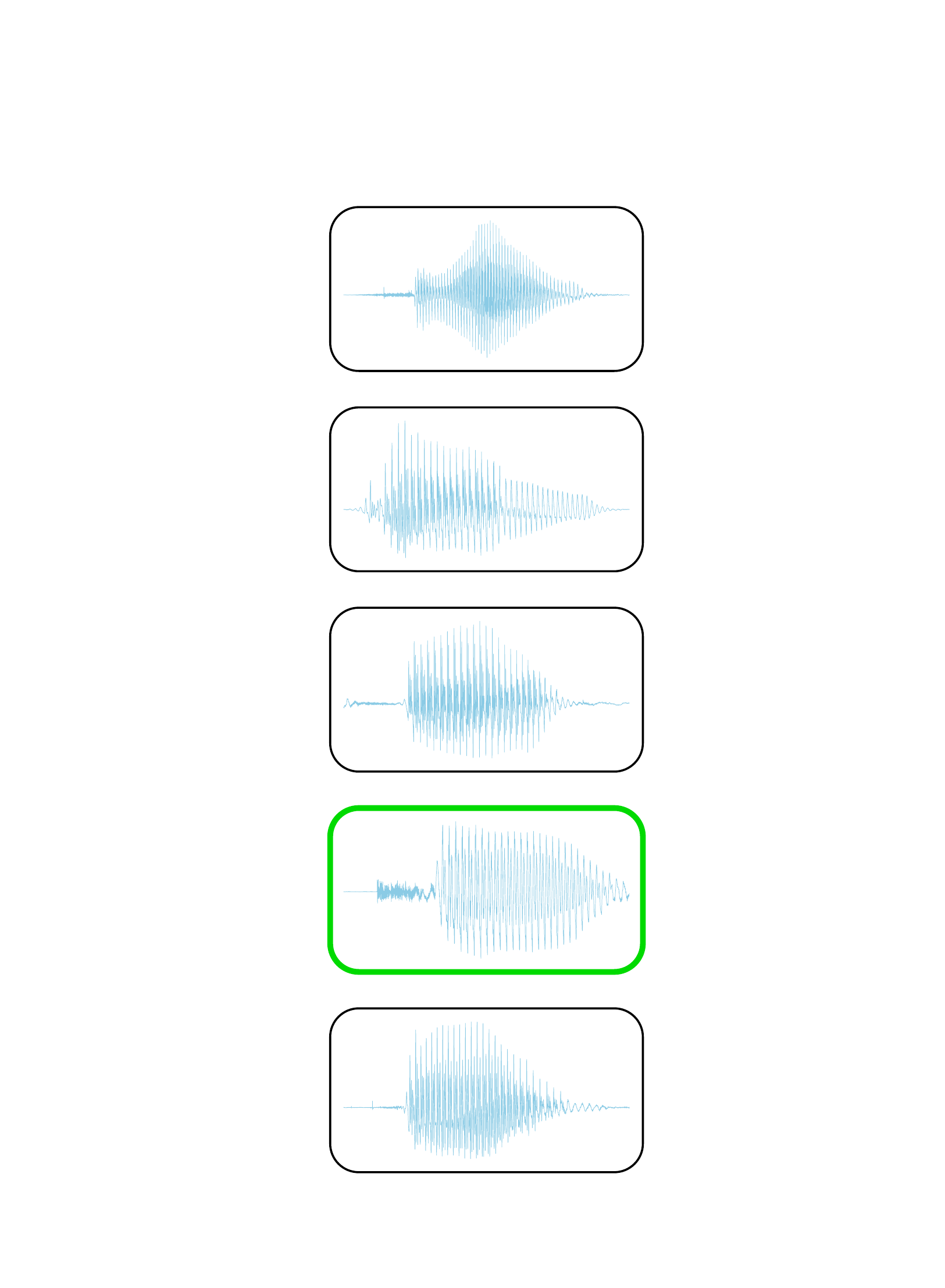
		}
		\vspace*{-0.2cm}
	\end{minipage}&
	\begin{minipage}[b]{.305\linewidth}  
		\centering
		\centerline{
			\footnotesize
			\def\svgscale{0.23}
\begingroup%
  \makeatletter%
  \providecommand\color[2][]{%
    \errmessage{(Inkscape) Color is used for the text in Inkscape, but the package 'color.sty' is not loaded}%
    \renewcommand\color[2][]{}%
  }%
  \providecommand\transparent[1]{%
    \errmessage{(Inkscape) Transparency is used (non-zero) for the text in Inkscape, but the package 'transparent.sty' is not loaded}%
    \renewcommand\transparent[1]{}%
  }%
  \providecommand\rotatebox[2]{#2}%
  \newcommand*\fsize{\dimexpr\f@size pt\relax}%
  \newcommand*\lineheight[1]{\fontsize{\fsize}{#1\fsize}\selectfont}%
  \ifx\svgwidth\undefined%
    \setlength{\unitlength}{680.31496063bp}%
    \ifx\svgscale\undefined%
      \relax%
    \else%
      \setlength{\unitlength}{\unitlength * \real{\svgscale}}%
    \fi%
  \else%
    \setlength{\unitlength}{\svgwidth}%
  \fi%
  \global\let\svgwidth\undefined%
  \global\let\svgscale\undefined%
  \makeatother%
  \begin{picture}(1,0.9375)%
    \lineheight{1}%
    \setlength\tabcolsep{0pt}%
    \put(0,0){\includegraphics[width=\unitlength,page=1]{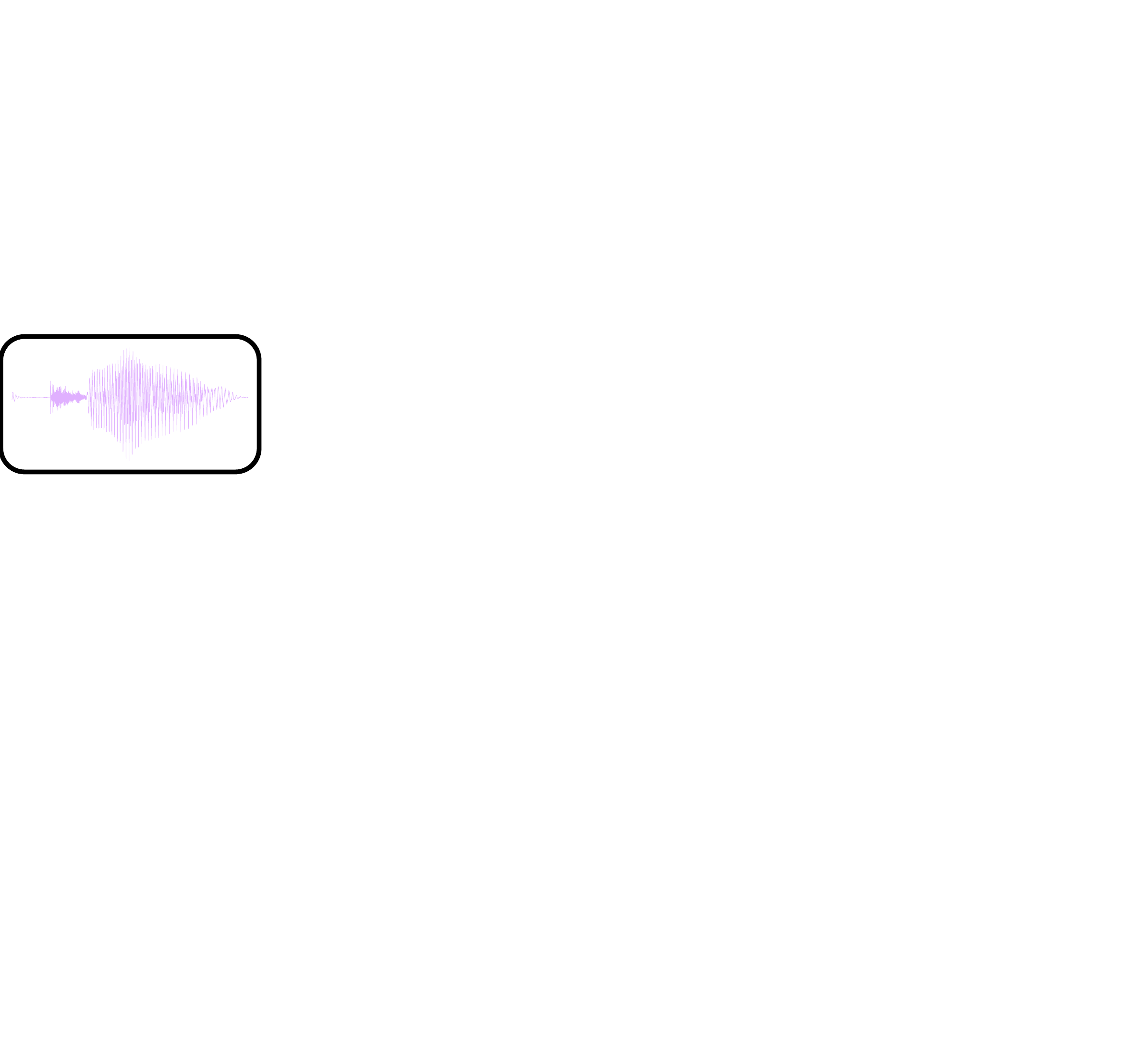}}%
    \put(-0.00151583,0.44037665){\color[rgb]{0,0,0}\makebox(0,0)[lt]{\lineheight{1.25}\smash{\begin{tabular}[t]{l}Query $\hat{\vx}_a$\\(\textit{``two''})\end{tabular}}}}%
    \put(0,0){\includegraphics[width=\unitlength,page=2]{cross_modal_task_infer.pdf}}%
    \put(0.14400312,0.73436664){\color[rgb]{0,0,0}\makebox(0,0)[lt]{\lineheight{1.25}\smash{\begin{tabular}[t]{l}$D_{\mathcal{S}}(\hat{\vx}_a, \hat{\vx}_v)$\end{tabular}}}}%
    \put(0,0){\includegraphics[width=\unitlength,page=3]{cross_modal_task_infer.pdf}}%
    \put(0.45616938,0.00519994){\color[rgb]{0,0,0}\makebox(0,0)[lt]{\lineheight{1.25}\smash{\begin{tabular}[t]{l}Matching set $\mathcal{M}_v$\\\end{tabular}}}}%
    \put(0.80208333,0.69846414){\color[rgb]{1,0,0}\makebox(0,0)[lt]{\lineheight{1.25}\smash{\begin{tabular}[t]{l}\textcolor{red}{Match}\end{tabular}}}}%
  \end{picture}%
\endgroup%

		}
		\vspace*{-0.2cm}
	\end{minipage}\\
	\mc{\vspace*{0.1cm}}&\mc{\vspace*{0.1cm}}\\
	\mc{\footnotesize \shortstack{i. One-shot multimodal \\ learning}}&
	\mc{\footnotesize \shortstack{ii. One-shot cross-modal \\ matching}}
\end{tabular}
}
\rmfamily  
\vspace*{-0.6\baselineskip}
\caption{\textup{(a)} Unimodal one-shot speech learning and classification and \textup{(b)} multimodal one-shot learning and matching of speech and images.
} 
\label{fig:fig_one_shot_tasks}
\vspace*{-0.4cm}
\end{figure*}

\section{Related Work}
\label{sec:related}

Most \hk{one-shot learning} studies have been primarily interested in image classification.
\hk{Apart from Siamese models~\cite{Koch2015Siamese}, which we focus on here}, other metric learning based approaches have been proposed~\cite{Vinyals2016MatchNet, Shyam2017AttenComparator, Snell2017ProtoNet} which build on advances in attention and memory mechanisms for neural networks.
Along with the more recent meta-learning approaches~\cite{Santoro2016MANN, Finn2017MAML, Mishra2018SNAIL}, these have each produced improvements in one-shot \hk{image classification}.
However, only small improvements have been made over Siamese networks: Finn {\it et al.}~\cite{Finn2017MAML} achieved state-of-the-art results with only 1.4\% increase in accuracy over Siamese networks for a 5-way one-shot learning task.

A limited number of studies have considered other domains, such as robotics~\cite{Walter2012OneShotRobot, Finn2017ImitationMAML},
video~\cite{Stafylakis2018ZeroShot},
and gesture recognition~\cite{Wu2012OneShot, Thomason2016RobotGestures}.
Lake \textit{et al.}~\cite{Lake2014OneShotSpeech} 
investigated one-shot speech learning 
\hk{using}
a generative \hk{hierarchical hidden Markov} model to recognise novel words from learned 
primitives.
This 
Bayesian model is based on prior work~\cite{Lake2013OneShotHB} \hk{in vision} and has displayed strong results,
\hk{although on visual tasks the Siamese model seems superior~\cite{Koch2015Siamese}}. 
Our work is also related to 
learning multimodal representations 
from paired images and unlabelled speech~\cite{Ngiam2011MultimodalDL, Harwath2016UnsupLearn, Leidel2017ModalInvar, Kamper2019UnsupSpeechVis, Kashyap2017MSc}.
We extend this research to the one-shot domain.

\section{Multimodal One-Shot Learning}
\label{sec:tasks}
\label{sec:one_shot_task}
The 
goal of \hk{unimodal} one-shot learning is to build a model that can acquire new concepts after \hk{observing} 
only a single \hk{labelled} example \hk{from each class}.
This model must then successfully generalise to new instances of those concepts in tasks such as classification or regression.
\hk{Formally, a model is shown a support set $\mathcal{S}$, containing one labelled example for each of $L$ classes.}
\hk{From this set,} it must learn a classifier $C_\mathcal{S}$ for unseen queries $\hat{\vx}$.
This 
\hk{is illustrated} 
in Figure~\ref{fig:fig_one_shot_tasks}(a) for five-way one-shot speech classification.
In this case the support set $\mathcal{S}$ contains spoken utterances 
along with hard textual labels. 
The model uses this information to classify the spoken \hk{test} query 
``two'' as the concept label \textit{two}. 
\hk{Note that the test-time query does not occur in the support set itself---it is an unseen instance of a class occurring in the support~set.}

We now modify this scheme to fit multimodal one-shot learning. 
Instead of a labelled unimodal support set, we are now given features in multiple modalities with the only supervisory signal being that these features co-occur.
In our case we consider speech and images as the two modalities, 
although this may be applied to any source of paired multi-sensory information.
\hk{Formally,} in an $L$-way one-shot problem 
we are given a multimodal support set 
$\mathcal{S} = \{ ( \vx_a^{(i)}, \vx_v^{(i)} ) \}_{i=1}^{L}$,
where each spoken caption $\vx_a^{(i)} \in \mathcal{A}$ (audio space) is paired with an image $\vx_v^{(i)} \in \mathcal{V}$ (vision space). 
\hk{During test-time, the modal is presented with a test query in one modality, and asked to determine the matching item in a test (or matching) set in the other modality.}
This is related to cross-modal retrieval tasks~\cite{Kashyap2017MSc, Wang2018CrossModal, Eisenschtat2017CrossModal} used to evaluate multimodal networks.
\hk{Formally,} 
we match query $\hat{\vx}_a$ in one modality (speech) to a matching set $\mathcal{M}_v = \{ ( \hat{\vx}_v ) \}_{i=1}^{N}$ 
in the other modality (images) according to some metric $D_\mathcal{S} (\hat{\vx}_a , \hat{\vx}_v)$ \hk{learned from the} support set $\mathcal{S}$. 
\hk{Neither the query $\hat{\vx}_a$ or the items in the match set $\mathcal{M}_v$ occur exactly in the support set $\mathcal{S}$.}
We refer to this task as one-shot cross-modal matching. 
It is illustrated in
Figure~\ref{fig:fig_one_shot_tasks}(b), where the spoken query ``two'' is most similar to the 
image of a \textit{two} \hk{in the matching set according} to the model trained on the support set.

\hk{One-shot learning can be generalised to $K$-shot learning, where, in the unimodal case,} a model is shown a support set containing $L$ novel classes and $K$ examples per class.
\hk{In multimodal} $L$-way $K$-shot \hk{learning}, the support set $\mathcal{S} = \{ ( \vx_a^{(i)}, \vx_v^{(i)} ) \}_{i=1}^{L \times K}$ \hk{consists of} $K$ speech-image example pairs for each of the $L$ classes.
\hk{This would occur, for instance, when a user teaches a robot speech-image correspondences by presenting it with multiple paired examples per~class.}


\section{Multimodal One-Shot Modelling}
\label{sec:cross_modal_matching}

\hk{We now turn to the general framework we use to perform multimodal one-shot learning.}
\hk{Assume we have a method or model that can measure similarity within a modality.}
One-shot cross-modal matching \hk{is then accomplished} by first comparing a query to all the items in the support set in the query modality (e.g.\ speech). 
\hk{The most similar (speech-image) support-set pair is retrieved.}
\hk{Finally, the retrieved support-set instance is used to determine the closest item in the matching set, based on comparisons in the matching-set modality (e.g.\ images).}
This approach \hk{thus} defines a metric $D_\mathcal{S}$ as a mapping $\mathcal{A} \rightarrow \mathcal{V}$ which can match speech to images by unimodal comparisons \hk{through}
the multimodal support set $\mathcal{S}$.
\hk{As a concrete example, in Figure~\ref{fig:fig_one_shot_tasks}(b) the speech query in (b)-ii is compared to all the support set speech segments in (b)-i, and the closest speech item determined.
The corresponding support-set image of this item 
is then taken and compared to all the images in the matching set $\mathcal{M}_v$ in (b)-ii. The closest image is predicted as the match.} 

\hk{Several different methods or models can be used to determine within-modality similarity:} we compare directly using the raw image pixels and extracted speech features (\S\ref{ssec:direct_features}) to feature embeddings learned by neural network functions (\S\ref{ssec:neural_classifier} and~\S\ref{ssec:siamese_network}).

\vspace*{-3pt}
\subsection{Direct feature matching}
\vspace*{-3pt}
\label{ssec:direct_features}
Our first approach consists of directly using image pixels and \hk{acoustic} speech features. 
\hk{We specifically use}
cosine similarity between image pixels and
dynamic time warping (DTW)~\cite{Sakoe1978DTW} 
\hk{to measure}
similarity between speech \hk{segments}. 
This is essentially our direct nearest neighbour baseline, as used in unimodal one-shot studies~\cite{Lake2014OneShotSpeech, Koch2015Siamese, Vinyals2016MatchNet}.

\vspace*{-3pt}
\subsection{Neural network classifiers}
\vspace*{-3pt}
\label{ssec:neural_classifier}

\hk{Another method, also used in unimodal one-shot learning, 
is to train a supervised model on a large background dataset. 
This background dataset should not contain instances of the target one-shot classes.
The idea is that features learned by such a model would still be useful for determining similarity on classes which it has not seen~\cite{Vinyals2016MatchNet}.
I.e., it follows the \textit{transfer learning} principle:}
first train a 
classifier on a large labelled dataset,
and then apply the learned representations to new tasks which are related but have too few training \hk{instances}~\cite{Donahue2014DeepFeats}.


\hk{Here we specifically}
train both a feedforward (FFNN) and a convolutional neural network (CNN) classifier. 
\hk{We train such networks separately for both the spoken and visual modalities.}
We then take representations from the final hidden layer (before the softmax \hk{output}) and apply these as the learned feature embeddings for nearest neighbour matching using cosine similarity.

\vspace*{-3pt}
\subsection{Siamese neural networks}
\vspace*{-3pt}
\label{ssec:siamese_network}

\hk{Networks can also be trained on background data to directly measure similarity between inputs instead of predicting a class label.}
Siamese neural networks have been used for this task~\cite{Bromley1993Siamese, Chopra2005Siamese}.
Two 
identical neural network functions with shared parameters (hence ``Siamese'') are trained to map input features to a target embedding space where the ``semantic'' \hk{relationship} 
between input pairs may be captured \hk{based on proximity: inputs of the same type should ideally be mapped to similar embeddings, while inputs that are unrelated should be far apart.}
\hk{Early approaches~\cite{Bromley1993Siamese, Chopra2005Siamese, Koch2015Siamese} took in pairs of training examples and either maximised or minimised a distance based on whether the inputs were of same or different types.}
Recent studies~\cite{Chechik2010OASIS, Wang2014TripNet, Hoffer2015TripletNet, Schroff2015FaceNet, hermann2014multilingual} have argued \hk{that the \textit{relative} rather than \textit{absolute} distance between embeddings are meaningful, and we also follow this approach here}.

\hk{Concretely, let $\vx_a$ and $\vx_p$ be inputs of the same class, while $\vx_a$ and $\vx_n$ are of different classes.}
The intuition is that \hk{we want} to push \hk{the so-called} anchor example $\vx_a$ \hk{and} positive example $\vx_p$ \hk{together such that the distance between them is smaller (by some specified margin) than the distance between the anchor $\vx_a$}
and negative example $\vx_n$.
\hk{Models using this approach are sometimes referred to as triplet models, since there are three tied networks for inputs} $(\vx_a, \vx_p, \vx_n)$.
We apply this 
approach where we learn embedding function $f(\cdot)$ as the final fully-connected layer of a CNN. 
We train models with a hinge loss for triplet pairs~\cite{Chechik2010OASIS, Wang2014TripNet, hermann2014multilingual, Schroff2015FaceNet}, defined as:
\vspace*{-3pt}
\begin{equation}
\label{eq_triplet_loss}
l(\vx_a, \vx_p, \vx_n) = \mathrm{max}\{ 0 , m + D(\vx_a, \vx_p) - D(\vx_a, \vx_n) \}
\end{equation}
where $D(\vx_1, \vx_2) = || f(\vx_1) - f(\vx_2) ||_2^2$ is \hk{the} squared Euclidean distance 
and $m$ is the margin 
between the pairs $(\vx_a, \vx_p)$ and $(\vx_a, \vx_n)$.

One problem with this approach is that the number of triplet pairs \hk{grows} cubically with the dataset,
and it may become infeasible to fit all possible triplets in memory.
We follow the \textit{online semi-hard mining} scheme~\cite{Schroff2015FaceNet},
where all possible anchor-positive pairs in a mini-batch are used.
{For each positive pair, the most difficult negative example $\vx_n$ satisfying $D(\vx_a, \vx_p) <  D(\vx_a, \vx_n)$ is then used, except if there is no such negative example in which case the one with the largest distance is used.}
According to~\cite{Schroff2015FaceNet}, although it might seem natural to simply choose the hardest negative example in the mini-batch, this constraint is required for stability.

\hk{We also incorporate another recent advance, specifically to improve efficience: we} 
simplify the three shared-parameter networks with a single neural network that embeds a mini-batch of examples and then samples triplet pairs online from these embeddings.
This is done with an efficient implementation of pairwise distances, similar to~\cite{Song2016LiftedStructLoss}.
We build this single network model with semi-hard triplet mining, and refer to it as \textit{Siamese CNN (online)}.
We also compare to using three shared-parameter networks with the same CNN architecture,
where we generate triplets offline at each training step from the current mini-batch.
We refer to this model as \textit{Siamese CNN (offline)} \hk{in our experiments}.
Similar to our neural network classifier baselines \hk{above}, these models are trained on triplets from a large disjoint labelled dataset \hk{which do not contain the target one-shot classes.} 

\section{Experiments}
\label{sec:experiment}
% ...

\vspace*{-3pt}
\subsection{Experimental setup}
\label{ssec:experiment_setup}
\vspace*{-3pt}

We \hk{perform} 
multimodal one-shot learning 
\hk{on}
a simple benchmark \hk{dataset}:
learning from 
examples of spoken digits \hk{paired} 
with handwritten digit images.
For speech we use the TIDigits corpus which contains spoken 
digit sequences from 326 different speakers~\cite{Leonard1993TIDigits}, 
and for images we use the MNIST handwritten digits dataset which contains 28$\times$28 grayscale images~\cite{LeCun1998MNIST}.
We use 
utterances
from men, women, and children, 
and split 
digit sequences into isolated digits using forced alignments.
\hk{Speech is parametrised as Mel-frequency cepstral coefficients with first and second order derivatives.} 
We centre \hk{zero-}pad or crop \hk{speech segments} 
to 120 frames. 
Image pixels are normalised to the range $[0, 1]$. 
\hk{Each isolated spoken digit is then paired with an image of the same type.}
Unlike previous work which used the same dataset combination for learning multimodal representations~\cite{Kashyap2017MSc, Leidel2017ModalInvar},
we treat utterances labelled ``oh'' and ``zero'' as separate classes, resulting in 11 \hk{class} labels.

Neural network models are trained on large labelled background datasets \hk{to obtain feature representations} 
which may be applied to the one-shot problem \hk{on classes not occurring in the background data}.
We use the 
\hk{speech} corpus of~\cite{Harwath2015FlickrAudio} and the
Omniglot handwritten character dataset~\cite{Lake2015Omniglot} as background data for the within-modality speech and vision models, respectively. 
Utterances in the audio corpus are split into isolated words using forced alignments, 
and 
features are extracted using the same process as for TIDigits.
\hk{We ensure that none of the target digit classes occur in this audio data.}
Images in the Omniglot dataset are downsampled to 28$\times$28 and pixel values are normalised and inverted in order to match MNIST.
\hk{Again, none of the Omniglot classes overlap with digit classes.}

Models are implemented in TensorFlow and trained using the Adam optimiser with a learning rate of $\textrm{10}^{-\textrm{3}}$ which is step decayed by 0.96 at each new epoch.
A batch size of 200 is used for the neural network classifiers (\S\ref{ssec:neural_classifier}). 
For the Siamese models (\S\ref{ssec:siamese_network}) we follow an alternate approach where mini-batches are formed using the \textit{batch all} strategy proposed in~\cite{Hermans2017TripletLoss}.
Specifically, we randomly sample $p$ classes and $k$ examples per class to produce balanced batches of $pk$ examples \hk{each}.
This results in $pk(pk-k)(k-1)$ valid triplet combinations, maximising the number of triplets within a mini-batch.
Our \textit{Siamese CNN (online)} variant is capable of large $pk$ combinations due to the efficient single network implementation and online triplet sampling scheme.
We 
choose $p=\textrm{128}$ and $k=\textrm{8}$ (total of 7\,282\,688 triplets per mini-batch).
\hk{For \textit{Siamese CNN (offline)}, trained in the standard way where three networks are explicitly tied~\cite{Wang2014TripNet, Hoffer2015TripletNet, Schroff2015FaceNet},}
\hk{we use}
$p=\textrm{32}$ and $k=\textrm{2}$, 
\hk{giving} 3\,968 triplets per mini-batch 
\hk{which is the largest batch we could fit on a single}
NVidia Titan Xp GPU.
Models are trained for a maximum of 100 epochs 
\hk{using} early stopping based on one-shot validation error on the background data.

\hk{We also tune all models using unimodal one-shot learning on the validation sets of the background data.}
\hk{This gave the following architecture for speech CNNs:} 39$\times$9 convolution with 128 filters; ReLU; 1$\times$3 max pooling; 1$\times$10 convolution with 128 filters; ReLU; 1$\times$28 max pooling over remaining units; 2048-unit fully-connected; ReLU.
\hk{Vision CNNs have the architecture:} 3$\times$3 convolution with 32 filters; ReLU; 2$\times$2 max pooling; 3$\times$3 convolution with 64 filters; ReLU; 2$\times$2 max pooling; 3$\times$3 convolution with 128 filters; ReLU; 2048-unit fully-connected; ReLU; 1024-unit fully-connected.
\hk{The speech and vision FFNNs have the same structure:} 3 fully-connected layers with 512 units each.
\hk{For the classifier networks (\S\ref{ssec:neural_classifier}), the speech networks have an additional 5534-unit softmax output layer (the number of word types in the background data), while vision networks have a 964-unit softmax (background image classes).}


We 
evaluate our models on the one-shot tasks according to the 
accuracy averaged over 400 test episodes.
Each 
\hk{episode} randomly samples a support set of isolated spoken digits paired with images 
for each of the $L = $11 
classes (the digits ``oh'' to ``nine'' and ``zero''). 
For testing, a matching set is sampled, containing 10 digit images not in the support set.
Finally, a random query is sampled, also not in the support set.
The query then needs to be matched to the correct item in the matching set.
\hk{The matching set only contains 10 items since there are 10 unique handwritten digit classes.} 
\hk{Within an episode, 10 different query instances are also sampled while keeping the support and matching sets fixed.}
Results
are averaged over 10 models trained with different seeds and we report average accuracies with 95\% confidence intervals.

\begin{table}[!t]
        \caption{11-way one-shot and five-shot speech classification results on isolated spoken digits. 
        }
        \vspace*{-7pt}
        \label{one-shot-speech-table}
        \begin{center}
                \setlength{\tabcolsep}{3.8pt}
                \begin{tabular*}{\linewidth}{l c c c}
                        \toprule
                        \multirow{2}{*}{\bf Model}      & {\bf Train}	& \multicolumn{2}{c}{\bf 11-way Accuracy} \\
             											& {\bf time}	& one-shot & five-shot \\
                        \midrule
                       	DTW						& --    & 67.99\% $\pm$ 0.29	& 91.30\% $\pm$ 0.20 \\
                        FFNN Classifier			& 13.1m	& 71.39\% $\pm$ 0.81	& 89.49\% $\pm$ 0.45 \\
                        CNN Classifier			& 60.6m	& 82.07\% $\pm$ 0.92	& 93.58\% $\pm$ 0.98 \\
                        \midrule
                        Siamese CNN (offline) 	& 70.5m & 89.40\% $\pm$ 0.54    & 95.12\% $\pm$ 0.37 \\
                        Siamese CNN (online) 	& 15.0m	& \bf 92.85\% $\pm$ 0.38& \bf 97.65\% $\pm$ 0.22 \\
                        \bottomrule
                \end{tabular*}
        \end{center}
        \vspace*{-0.7cm}
\end{table}

\vspace*{-3pt}
\subsection{One-shot speech classification}
\vspace*{-3pt}
\label{ssec:experiment_unimodal}

\hk{We first consider unimodal one-shot speech classification, which has so far only been considered in~\cite{Lake2014OneShotSpeech}.}
Table~\ref{one-shot-speech-table} shows one-shot and five-shot \hk{(see end of \S\ref{sec:one_shot_task})} speech classification results. 
Average training time is also shown; all models trained within a few seconds of the average. 
Siamese models outperform the 
direct feature matching baseline using DTW, as well as the 
neural network classifiers. 
\hk{The \textit{Siamese CNN (online)} model achieves best overall performance, outperforming the} \textit{Siamese CNN (offline)} variant, while training almost five times faster. 
The single network with online semi-hard mining is thus more efficient and accurate than the three shared-network approach.
\hk{None of these Siamese models were considered in~\cite{Lake2014OneShotSpeech}.}

\vspace*{-3pt}
\subsection{One-shot cross-modal matching of speech to images}
\vspace*{-3pt}
\label{ssec:experiment_cross_modal}

\hk{We now turn to}
%
multimodal one-shot learning.
Table~\ref{cross-modal-table} shows 
results for one- and five-shot cross-modal matching of speech to images.
\hk{Here}
the Siamese models are \hk{again} stronger overall compared to direct feature matching 
or transferring features from 
neural network classifiers.
The \textit{Siamese CNN (online)} model achieves our best 
results, 
with double the 
accuracy of direct feature matching using pixel-distance over images and DTW over speech.
The \textit{Siamese CNN (offline)} model follows closely\hk{, but is again slower to train}.

While the Siamese models achieve promising results compared to the baselines here, our best one-shot multimodal accuracy is 
\hk{lower than the accuracy in unimodal one-shot speech classification (Table~\ref{one-shot-speech-table}).}
\hk{Our best unimodal one-shot vision model achieves an accuracy of 74\% 
(comparing favourably to the best result of 72\% reported in~\cite{Vinyals2016MatchNet}).}
\hk{The multimodal one-shot results here are therefore worse than both the individual unimodal matching results.}
This \hk{is due to compounding errors in our retrieval framework (\S\ref{sec:cross_modal_matching}): }
errors in 
comparisons with the support set affects comparisons in the \hk{subsequent matching step.} 
This suggests investigating an end-to-end architecture which can directly compare test queries \hk{in one modality} to the matching set items in the other modality, \hk{without doing explicit comparisons to the support set}.

\begin{table}[!t]
	\caption{11-way one- and five-shot cross-modal matching of spoken and visual digits.}
	\vspace*{-7pt}
	\label{cross-modal-table}
	\begin{center}
		\begin{tabular}{l c c}
			\toprule
			\multirow{2}{*}{\bf Model}	& \multicolumn{2}{c}{\bf 11-way Accuracy} \\
										& one-shot & five-shot \\
			\midrule
			DTW + Pixels				& 34.92\% $\pm$ 0.42	& 44.46\% $\pm$ 0.69 \\
			FFNN Classifier				& 36.49\% $\pm$ 0.41	& 44.29\% $\pm$ 0.56 \\
			CNN Classifier				& 56.47\% $\pm$ 0.76	& 63.97\% $\pm$ 0.91 \\
			\midrule
			Siamese CNN (offline)		& 67.41\% $\pm$ 0.56	& 70.92\% $\pm$ 0.36 \\
			 Siamese CNN (online)	        & \bf 70.12\% $\pm$ 0.68	& \bf 73.53\% $\pm$ 0.52 \\
			\bottomrule
		\end{tabular}
	\end{center}
	\vspace*{-0.8cm}
\end{table}

\subsection{Invariance to speakers}
\vspace*{-3pt}
\label{ssec:experiment_speaker_invariance}
In all the experiments above we chose spoken queries such that the speaker \hk{uttering the query} does not appear in the support set.
\hk{This is representative of an extreme case where one user teaches an agent and another then uses the system.}
\hk{An even more extreme case could occur: the matching item in the support set could be the only item not coming from the query speaker.
This is problematic since the same word uttered by different speakers might be acoustically more different than different words uttered by the same speaker.}
We test this worst-case setting:
we sample a 
support set 
where all spoken 
digits are from the same speaker as the query,
except for the one instance matching the query word which is produced by a different speaker. 
Cross-modal matching results 
\hk{for this case} are shown in Table~\ref{cross-modal-invariance-table}.
All of the models experience a \hk{drop} in accuracy compared to the results in Table~\ref{cross-modal-table} \hk{(first column)}.
\hk{This decrease is smallest for the Siamese models, with the \textit{DTW + Pixels} approach dropping most.}
\hk{This indicates that the}
neural models learn features \hk{from the background data} which are more independent of speaker and can generalise to other speakers,
whereas DTW over speech \hk{is affected more by speaker mismatch.}

\begin{table}[!t]
	\caption{Speaker invariance tests for 11-way one-shot cross-modal speech-image digit matching. All 
	support set items are from the same speaker as the query, 
	except for the 
	item actually matching the~query.}
	\label{cross-modal-invariance-table}
	\vspace*{-7pt}
	\begin{center}
		\begin{tabular}{l c}
			\toprule
			\multirow{2}{*}{\bf Model}	& \multicolumn{1}{c}{\bf 11-way Accuracy} \\
										& one-shot \\
			\midrule
			DTW + Pixels				& 28.00\% $\pm$ 1.86 \\
			FFNN Classifier				& 34.95\% $\pm$ 2.28 \\
			CNN Classifier				& 53.71\% $\pm$ 2.20 \\
			\midrule
			Siamese CNN (offline)		& 66.70\% $\pm$ 0.92 \\
			 Siamese CNN (online)	& \bf 69.73\% $\pm$ 1.04 \\
			\bottomrule
		\end{tabular}
	\end{center}
	\vspace*{-0.7cm}
\end{table}


\section{Conclusion}
\vspace*{-0.1cm}
\label{sec:conclusion}

We introduced and formalised multimodal one-shot learning, specifically for learning from spoken and visual representations of digits.
Observing just one paired example from each class, a model is asked to pick the correct digit image for an unseen spoken query.
We proposed and evaluated several baseline and more advanced models.
Although our Siamese convolutional approach outperforms a raw-feature nearest neighbour model, the performance in the cross-modal case is still worse than in unimodal one-shot learning.
We argued that this is due to a compounding of errors in our framework, which relies on successive unimodal comparisons.
In future work, we will therefore explore a method that can directly match one modality to another, particularly looking into recent meta-learning approaches.\blfootnote{We thank NVIDIA for sponsoring a Titan Xp GPU for this work.}

\vfill

\pagebreak
\bibliographystyle{IEEEbib_modified}
\bibliography{icassp2019}

\begin{thebibliography}{10}

\bibitem{Halberda2006Dax}
J.~Halberda,
\newblock ``Is this a dax which {I} see before me? {U}se of the logical
  argument disjunctive syllogism supports word-learning in children and
  adults,''
\newblock {\em Cogn. Psychol.}, vol. 53, no. 4, pp. 310--344, 2006.

\bibitem{FeiFei2006OneShot}
L.~Fei-Fei, R.~Fergus, and P.~Perona,
\newblock ``One-shot learning of object categories,''
\newblock {\em IEEE Trans. PAMI}, vol. 28, no. 4, pp. 594--611, 2006.

\bibitem{Lake2011OneShotVis}
B.~M. Lake, R.~Salakhutdinov, J.~Gross, and J.~B. Tenenbaum,
\newblock ``One shot learning of simple visual concepts,''
\newblock in {\em Proc. CogSci}, 2011.

\bibitem{Koch2015Siamese}
G.~Koch, R.~Zemel, and R.~Salakhutdinov,
\newblock ``Siamese neural networks for one-shot image recognition,''
\newblock in {\em Proc. ICML}, 2015.

\bibitem{Rasanen2015HumanLang}
O.~R{\"a}s{\"a}nen and H.~Rasilo,
\newblock ``A joint model of word segmentation and meaning acquisition through
  cross-situational learning,''
\newblock {\em Psychol. Rev.}, vol. 122, no. 4, pp. 792--829, 2015.

\bibitem{Besacier2014LowResASR}
L.~Besacier, E.~Barnard, A.~Karpov, and T.~Schultz,
\newblock ``Automatic speech recognition for under-resourced languages: A
  survey,''
\newblock {\em Speech Commun.}, vol. 56, pp. 85--100, 2014.

\bibitem{Walter2012OneShotRobot}
M.~R. Walter, Y.~Friedman, M.~Anton, and S.~Teller,
\newblock ``One-shot visual appearance learning for mobile manipulation,''
\newblock {\em IJRR}, vol. 31, no. 4, pp. 554--567, 2012.

\bibitem{Taniguchi2016Robotics}
T.~Taniguchi et~al.,
\newblock ``Symbol emergence in robotics: A survey,''
\newblock {\em Adv. Robot.}, vol. 30, no. 11--12, pp. 706--728, 2016.

\bibitem{Thomason2016RobotGestures}
W.~Thomason and R.~A. Knepper,
\newblock ``Recognizing unfamiliar gestures for human-robot interaction through
  zero-shot learning,''
\newblock in {\em Proc. ISER}, 2016.

\bibitem{Renkens2017SpeechAcqui}
V.~Renkens and H.~{Van hamme},
\newblock ``Weakly supervised learning of hidden {M}arkov models for spoken
  language acquisition,''
\newblock {\em IEEE/ACM Trans. ASLP}, vol. 25, no. 2, pp. 285--295, 2017.

\bibitem{Renkens2018CapsNetSLU}
V.~Renkens and H.~{Van hamme},
\newblock ``Capsule networks for low resource spoken language understanding,''
\newblock in {\em Proc. Interspeech}, 2018.

\bibitem{Bromley1993Siamese}
J.~Bromley et~al.,
\newblock ``Signature verification using a ``{Siamese}'' time delay neural
  network,''
\newblock in {\em Proc. NIPS}, 1994.

\bibitem{Chopra2005Siamese}
S.~Chopra, R.~Hadsell, and Y.~LeCun,
\newblock ``Learning a similarity metric discriminatively, with application to
  face verification,''
\newblock in {\em Proc. CVPR}, 2005.

\bibitem{Chechik2010OASIS}
G.~Chechik, V.~Sharma, U.~Shalit, and S.~Bengio,
\newblock ``Large scale online learning of image similarity through ranking,''
\newblock {\em JMLR}, vol. 11, no. 11, pp. 1109--1135, 2010.

\bibitem{Wang2014TripNet}
J.~Wang et~al.,
\newblock ``Learning fine-grained image similarity with deep ranking,''
\newblock in {\em Proc. CVPR}, 2014.

\bibitem{hermann2014multilingual}
K.~M. Hermann and P.~Blunsom,
\newblock ``Multilingual distributed representations without word alignment,''
\newblock in {\em Proc. ICLR}, 2014.

\bibitem{Hoffer2015TripletNet}
E.~Hoffer and N.~Ailon,
\newblock ``Deep metric learning using triplet network,''
\newblock in {\em Proc. SIMBAD}, 2015.

\bibitem{Schroff2015FaceNet}
F.~Schroff, D.~Kalenichenko, and J.~Philbin,
\newblock ``Facenet: A unified embedding for face recognition and clustering,''
\newblock in {\em Proc. CVPR}, 2015.

\bibitem{Hermans2017TripletLoss}
A.~Hermans, L.~Beyer, and B.~Leibe,
\newblock ``In defense of the triplet loss for person re-identification,''
\newblock {\em arXiv preprint arXiv:1703.07737}, 2017.

\bibitem{Lake2014OneShotSpeech}
B.~M. Lake, C.-Y. Lee, J.~R. Glass, and J.~B. Tenenbaum,
\newblock ``One-shot learning of generative speech concepts,''
\newblock in {\em Proc. CogSci}, 2014.

\bibitem{Vinyals2016MatchNet}
O.~Vinyals et~al.,
\newblock ``Matching networks for one shot learning,''
\newblock in {\em Proc. NIPS}, 2016.

\bibitem{Shyam2017AttenComparator}
P.~Shyam, S.~Gupta, and A.~Dukkipati,
\newblock ``Attentive recurrent comparators,''
\newblock in {\em Proc. ICML}, 2017.

\bibitem{Snell2017ProtoNet}
J.~Snell, K.~Swersky, and R.~Zemel,
\newblock ``Prototypical networks for few-shot learning,''
\newblock in {\em Proc. NIPS}, 2017.

\bibitem{Santoro2016MANN}
A.~Santoro et~al.,
\newblock ``Meta-learning with memory-augmented neural networks,''
\newblock in {\em Proc. ICML}, 2016.

\bibitem{Finn2017MAML}
C.~Finn, P.~Abbeel, and S.~Levine,
\newblock ``Model-agnostic meta-learning for fast adaptation of deep
  networks,''
\newblock in {\em Proc. ICML}, 2017.

\bibitem{Mishra2018SNAIL}
N.~Mishra, M.~Rohaninejad, X.~Chen, and P.~Abbeel,
\newblock ``A simple neural attentive meta-learner,''
\newblock in {\em Proc. ICLR}, 2018.

\bibitem{Finn2017ImitationMAML}
C.~Finn et~al.,
\newblock ``One-shot visual imitation learning via meta-learning,''
\newblock in {\em Proc. Robot Learn.}, 2017.

\bibitem{Stafylakis2018ZeroShot}
T.~Stafylakis and G.~Tzimiropoulos,
\newblock ``Zero-shot keyword spotting for visual speech recognition
  in-the-wild,''
\newblock in {\em Proc. ECCV}, 2018.

\bibitem{Wu2012OneShot}
D.~Wu, F.~Zhu, and L.~Shao,
\newblock ``One shot learning gesture recognition from rgbd images,''
\newblock in {\em Proc. CVPR}, 2012.

\bibitem{Lake2013OneShotHB}
B.~M. Lake, R.~Salakhutdinov, and J.~B. Tenenbaum,
\newblock ``One-shot learning by inverting a compositional causal process,''
\newblock in {\em Proc. NIPS}, 2013.

\bibitem{Ngiam2011MultimodalDL}
J.~Ngiam et~al.,
\newblock ``Multimodal deep learning,''
\newblock in {\em Proc. ICML}, 2011.

\bibitem{Harwath2016UnsupLearn}
D.~Harwath, A.~Torralba, and J.~R. Glass,
\newblock ``Unsupervised learning of spoken language with visual context,''
\newblock in {\em Proc. NIPS}, 2016.

\bibitem{Leidel2017ModalInvar}
K.~Leidal, D.~Harwath, and J.~R. Glass,
\newblock ``Learning modality-invariant representations for speech and
  images,''
\newblock in {\em Proc. ASRU}, 2017.

\bibitem{Kamper2019UnsupSpeechVis}
H.~Kamper, G.~Shakhnarovich, and K.~Livescu,
\newblock ``Semantic speech retrieval with a visually grounded model of
  untranscribed speech,''
\newblock {\em IEEE/ACM Trans. ASLP}, vol. 27, no. 1, pp. 89--98, 2019.

\bibitem{Kashyap2017MSc}
K.~Kashyap,
\newblock ``Learning digits via joint audio-visual representations,''
\newblock M.S. thesis, MIT, 2017.

\bibitem{Wang2018CrossModal}
L.~Wang, Y.~Li, J.~Huang, and S.~Lazebnik,
\newblock ``Learning two-branch neural networks for image-text matching
  tasks,''
\newblock {\em IEEE Trans. PAMI}, vol. 41, no. 2, pp. 394--407, 2018.

\bibitem{Eisenschtat2017CrossModal}
A.~Eisenschtat and L.~Wolf,
\newblock ``Linking image and text with 2-way nets,''
\newblock in {\em Proc. CVPR}, 2017.

\bibitem{Sakoe1978DTW}
H.~Sakoe and S.~Chiba,
\newblock ``Dynamic programming algorithm optimization for spoken word
  recognition,''
\newblock {\em IEEE Trans. ASSP}, vol. 26, no. 1, pp. 43--49, 1978.

\bibitem{Donahue2014DeepFeats}
J.~Donahue et~al.,
\newblock ``Decaf: A deep convolutional activation feature for generic visual
  recognition,''
\newblock in {\em Proc. ICML}, 2014.

\bibitem{Song2016LiftedStructLoss}
H.~O. Song, Y.~Xiang, S.~Jegelka, and S.~Savarese,
\newblock ``Deep metric learning via lifted structured feature embedding,''
\newblock in {\em Proc. CVPR}, 2016.

\bibitem{Leonard1993TIDigits}
R.~G. Leonard and G.~Doddington,
\newblock ``{TIDIGITS LDC93S10},'' Philadelphia: Linguistic Data Consortium,
  1993.

\bibitem{LeCun1998MNIST}
Y.~Le{C}un, L.~Bottou, Y.~Bengio, and P.~Haffner,
\newblock ``Gradient-based learning applied to document recognition,''
\newblock {\em Proc. IEEE}, vol. 86, no. 11, pp. 2278--2324, 1998.

\bibitem{Harwath2015FlickrAudio}
D.~Harwath and J.~R. Glass,
\newblock ``Deep multimodal semantic embeddings for speech and images,''
\newblock in {\em Proc. ASRU}, 2015.

\bibitem{Lake2015Omniglot}
B.~M. Lake, R.~Salakhutdinov, and J.~B. Tenenbaum,
\newblock ``Human-level concept learning through probabilistic program
  induction,''
\newblock {\em Science}, vol. 350, no. 6266, pp. 1332--1338, 2015.

\end{thebibliography}

\end{document}